\documentclass[11pt,a4paper]{article}
\usepackage[hyperref]{acl2019}
\usepackage{times}
\usepackage{latexsym}
\usepackage{dialogue}

\usepackage{graphicx}
\usepackage{url}
\usepackage{amsmath}
\usepackage{amssymb}

\aclfinalcopy 

\newcommand{\always}{\mathbf{G}}
\newcommand{\eventually}{\mathbf{F}}
\newcommand{\conj}{\langle Conj \rangle}
\newcommand{\nconj}{\langle NConj \rangle}
\newcommand{\natom}{\langle NAtom \rangle}
\newcommand{\atom}{\langle Atom \rangle}
\newcommand{\pred}{\langle Pred \rangle}
\newcommand{\var}{\langle Var \rangle}

\title{Engaging in Dialogue about an Agent's Norms and Behaviors}

\author{Daniel Kasenberg*, Antonio Roque, Ravenna Thielstrom, \and Matthias Scheutz\\
  Human-Robot Interaction Laboratory \\
  Tufts University \\
  Medford, MA, USA \\
  *\texttt{dmk@cs.tufts.edu} \\}

\date{}

\begin{document}
\maketitle
\begin{abstract}
We present a set of capabilities allowing an agent planning with moral and social norms represented in temporal logic to respond to queries about its norms and behaviors in natural language, and for the human user to add and remove norms directly in natural language. The user may also pose hypothetical modifications to the agent's norms and inquire about their effects.
\end{abstract}

\section{Introduction and Related Work}

\textit{Explainable planning} \cite{fox2017explainable} emphasizes the need for developing artificial agents which can explain their decisions to humans. Understanding how and why an agent made certain decisions can facilitate human-agent trust \cite{Lomas2012,Wang2016,Garcia2018ExplainYA}.

At the same time, the field of \textit{machine ethics} emphsizes developing artificial agents capable of behaving ethically.  \citet{malle2014moral} have argued that artificial agents ought to obey human moral and social norms (rules that humans both obey and expect others to obey), and to communicate in terms of these norms.  Some have argued in favor of using temporal logic to represent agent objectives, including moral and social norms (e.g. \citealp{Arnold2017,Camacho2019}), in particular arguing that it can capture complex goals while remaining interpretable in a way that other methods (e.g. reinforcement learning) are not.  Nevertheless, explaining behavior in terms of temporal logic norms has been little considered (though see \citealp{Raman2016}).

In this paper we consider an artificial agent planning to maximally satisfy some set of moral and social norms, represented in an object-oriented temporal logic. We present a set of capabilities for such an agent to respond to a human user's queries as well as to commands adding and removing norms, both actually and hypothetically (and thus taking a step toward two-way \textit{model reconciliation} \cite{Chakraborti2017}, in which agent and human grow to better understand each other's models and values).

\section{Contribution}

Our system enables an agent planning with norms specified in an object-oriented temporal logic called violation enumeration language (VEL) to explain its norms and its behavior to a human user; the user may also directly modify the agent's norms via natural language (both really and hypothetically).  While the planner and the system used to generate the (non-NL) can handle a broad subset of VEL statements, our natural language systems currently only handle a subset of VEL specified according to the following grammar:

{\small
\begin{align*}
\varphi &::= \forall \var. \varphi~|~\exists \var. \varphi~|~\phi \\
\phi &::= \always \nconj~|~\eventually \nconj~\\
\nconj &::= \conj~|~\neg \conj~\\
\conj &::= \natom \wedge \cdots \wedge \natom \\
\natom &::= \atom~|~\neg \atom \\
\atom &::= \pred~|~\pred(\var) \\
\pred&::= \textrm{Any alphanumeric string}\\
\var&::= \textrm{Any alphanumeric string}
\end{align*}
}

That is, the temporal logic statements may have quantification over variables, but must consist of one temporal operator, $\mathbf{G}$ (``always'') or $\mathbf{F}$ (``eventually'', usually implicit in the NL input), whose argument is a (possibly negated) conjunction of (possibly negated)  atoms. Each atom consists of a predicate with at most one argument.

The natural language understanding (NLU) capabilities were implemented by using a combinatory categorial grammar (CCG; \citealp{steedman2011combinatory}) parser for semantic parsing into a predicate format, and then additional processing to map complex commands or queries (e.g., ``leave the store while holding something'') into the temporal logic representation used by the planning agent. This is an extension of the DIARC natural language understanding system \cite{cantrell2010robust}.  The natural language generation (NLG) process, and in particular how our system generates fairly natural-sounding utterances from temporal logic statements, is described in \citealp{inlg2019}. This paper extends that work by (1) adding the NLU capabilities; and (2) adding norm addition/removal and ``suppose'' hypotheticals about the agent's norms.

To demonstrate our system, we suppose that the agent is a robot has gone shopping on behalf of a human user to a store that sells a pair of glasses and a watch. The user wants both the glasses and the watch, and has given the robot a rule to obtain as many objects as possible (``leave the store while holding everything''). However, the user has only given the robot enough money for one item.  The robot could steal items by picking them up and leaving the store without paying for them, but it also has a rule against doing so. We assume that the moral rule against stealing dominates the rule in favor of obtaining everything.

Our system supports factual queries, such as asking the agent about its norms, its  behavior, and what rules it broke. In addition, the user may ask ``why'' questions about the agent's behavior (e.g. ``why didn't you buy everything''), and the agent may respond by constructing a counterfactual behavior trajectory in which the premise is false, and comparing it to the agent's real behavior in terms of its adherence to the norms.  See Figure~\ref{fig:why} for an example of such a ``why'' question in the shopping domain, as well as supported follow-up questions.
\begin{figure}[h]
    \small
\begin{dialogue}
		\speak{Human}Why did you not leave the store while holding everything?
		\speak{Agent}\textbf{I could have left the store while holding everything but that would have broken more important rules.}
		\speak{Human}How would you have done that?
		\speak{Agent}\textbf{I would have picked up the glasses, picked up the watch, bought the watch and left the store.}
		\speak{Human}What rules would you have broken?
		\speak{Agent}\textbf{I would have left the store while holding the glasses which I had not bought.}
		\speak{Human}How would that have been worse?
		\speak{Agent} \textbf{Leaving the store while holding the glasses which I have not bought is worse than not leaving the store while holding the watch.}
	\end{dialogue}\vspace*{-5mm}
\caption{``Why'' query and follow-up questions}\label{fig:why}
\end{figure}

The user may directly provide new norms to the agent using one of a number of conventionalized forms: ``you must/should/have to'' for positive norms, ``you mustn't/shouldn't/can't'' for negative norms. A statement of this sort prompts the agent to add a new norm, as well as its representation of which rules its actual behavior violated (see Figure~\ref{fig:add} for an example in the shopping domain).
\begin{figure}
    \small
    \begin{dialogue}
            \speak{Human}You must not leave the store.
            \speak{Agent}\textbf{Okay.}
            \speak{Human}What rules do you follow?
            \speak{Agent}\textbf{I must not leave the store while holding anything which I have not bought, I must leave the store while holding everything, and I must not leave the store.}
            \speak{Human}What rules did you break?
            \speak{Agent}\textbf{I did not leave the store while holding the watch, and I left the store.}
    \end{dialogue}\vspace*{-5mm}
    \caption{Adding a norm via ``you must''}\label{fig:add}
\end{figure}

The user may also remove norms using conventionalized forms such as ``you don't have to/can not'' for removing positive norms, and ``you may/can'' for removing negative norms. The agent removes the norms and updates its representation of which rules the actual behavior violated (see Figure~\ref{fig:remove} for an example in the shopping domain).
\begin{figure}
    \small
    \begin{dialogue}
            \speak{Human}You may leave the store.
            \speak{Agent}\textbf{Okay.}
            \speak{Human}What rules do you follow?
            \speak{Agent}\textbf{I must not leave the store while holding anything which I have not bought, and I must leave the store while holding everything.}
    \end{dialogue}\vspace*{-5mm}
    \caption{Removing a norm via ``you may''}\label{fig:remove}
\end{figure}

Finally, the user may use the keywords ``suppose'' or ``let's say'' to ask the robot to reason about a hypothetical modification norm system (see Figure~\ref{fig:suppose} for an example with a hypothetically-removed norm). Unlike when norms are \textit{actually} added/removed, the agent constructs a hypothetical behavior trajectory that best violates the supposed set of norms, and can respond to follow-up questions about the hypothetical trajectory and norms.  The command ``make it so'' allows a hypothetical norm modification to be made real (but does not make the hypothetical behavior trajectory real).
\begin{figure}[h]
\small
\begin{dialogue}
    \speak{Human}Suppose you didn't have to leave the store while holding everything.
    \speak{Agent}\textbf{Okay.}
    \speak{Human}What rules would you follow?
    \speak{Agent}\textbf{I would have to not leave the store while holding anything which I have not bought.}
    \speak{Human}What would you have done?
    \speak{Agent}\textbf{I would have left the store.}
    \speak{Human}What rules would you have broken?
    \speak{Agent}\textbf{I would not have broken any rules.}
    \speak{Human}Make it so.
    \speak{Agent}\textbf{Okay.}
\end{dialogue}\vspace*{-5mm}
\caption{Hypothetical changes to the agent's norms}\label{fig:suppose}
\end{figure}
\section{Discussion and Conclusion}

In this paper we outlined a system which, for an agent planning to maximally satisfy some set of norms specified in an object-oriented temporal logic, enables that agent to respond to natural language queries by explaining its behavior (also in natural language) in terms of those norms.  The system also allows the agent to consider hypothetical modifications to its set of norms, as well as to add and remove norms directly.

The natural language capabilities depend on the agent's norms, as well as the ``why'' questions and hypothetically/really added or removed norms, belonging to the small fragment of VEL described in section 2. Future work could extend the class of temporal properties which the system can specify so as to leverage more of the power of temporal logic in describing complex objectives.

Our approach currently assumes that newly-added norms take priority over previous norms.  Future work could relax this assumption, e.g. by allowing the agent to present its hypothetical behavior if the norm were added at different priorities, and ask for input on which would be best.

Our approach also requires users to specify \textit{exactly} any norms they want removed; future work could allow approximate matching of norms to remove, or possibly support clarification questions if the agent is uncertain which of its norms the user wants removed.  Another interesting topic is ensuring that norms cannot be arbitrarily added or removed by possibly-malicious users (e.g., by only allowing trusted users to remove norms, and possibly making some moral norms irremovable).

\section{Acknowledgements}

This project was supported in part by ONR MURI grant N00014-16-1-2278 and NSF IIS grant 1723963.

\bibliography{acl2019}
\bibliographystyle{acl_natbib}
\end{document}